%% file: output.tex
\pdfoutput=1

\documentclass[11pt]{article}

\usepackage{EMNLP2023}

\usepackage{times}
\usepackage{latexsym}
\usepackage{multirow}
\usepackage{flushend}

\usepackage[colorinlistoftodos]{todonotes}
\usepackage[T1]{fontenc}

\usepackage[utf8]{inputenc}

\usepackage{microtype}

\usepackage{inconsolata}
\usepackage{graphicx}
\usepackage{amsmath}
\usepackage{hyperref}
\usepackage{cleveref}
\usepackage{algpseudocode}
\usepackage{algorithm}

\usepackage[most]{tcolorbox}

\usepackage{listings}
\usepackage{hyperref}  
\usepackage[acronym,nohypertypes={acronym}]{glossaries}
\usepackage{sankey}
\usepackage{pgfplots}
\usepackage{pgfplotstable}

\usepackage{pgfplots}
\pgfplotsset{compat=1.16} 
\usepgfplotslibrary{external}
\algnewcommand{\Inputs}[1]{%
  \State \textbf{Inputs:}
  \Statex \hspace*{\algorithmicindent}\parbox[t]{.8\linewidth}{\raggedright #1}
}
\algnewcommand{\Initialize}[1]{%
  \State \textbf{Initialize:}
  \Statex \hspace*{\algorithmicindent}\parbox[t]{.8\linewidth}{\raggedright #1}
}

\usepackage{soul}
\definecolor{green}{rgb}{0.0, 0.4, 0.0}
\definecolor{purple}{rgb}{0.5, 0.2, 0.8}

\input{acronyms}

\title{Automated test generation to evaluate tool-augmented LLMs 
\\ as conversational AI agents}

\author{Samuel Arcadinho \\
  \texttt{samuel.arcadinho@zendesk.com} \\\And
  David Aparício \\
  \texttt{david.aparicio@zendesk.com} \\\AND
  Mariana S. C. Almeida \\
  \texttt{mariana.almeida@zendesk.com} \\}

\begin{document}

\maketitle
\begin{abstract}
Tool-augmented LLMs are a promising approach to create AI agents that can have realistic conversations, follow procedures, and call appropriate functions. However, evaluating them is challenging due to the diversity of possible conversations, and existing datasets focus only on single interactions and function-calling. We present a test generation pipeline to evaluate LLMs as conversational AI agents. Our framework uses LLMs to generate diverse 
tests grounded on user-defined procedures. For that, we use intermediate graphs to limit the LLM test generator's tendency to hallucinate content that is
not grounded on input procedures, and enforces high coverage of the possible conversations. Additionally, we put forward \texttt{ALMITA}, a manually curated dataset for evaluating AI agents in customer support, and use it to evaluate existing LLMs. Our results show that while tool-augmented LLMs perform well in single interactions, they often struggle to handle complete conversations. While our focus is on customer support, our method is general and capable of AI agents for different domains.

\end{abstract}

\section{Introduction}


\Glspl{llm} are revolutionizing AI agents and have demonstrated remarkable generalization capabilities across various domains \cite{wu2023autogen, lan2024teachers, li2024personal}. In particular, \Glspl{llm} have made a profound impact as chatbots and as AI agents in customer support systems \cite{dam2024complete, katragadda2024leveraging}. 

Nevertheless, carelessly deploying an LLM as an AI agent, and allowing them to interact with real users and APIs, can lead to misinformation, reputational damage and costs to the company. Thus, it is critical to evaluate AI agents beforehand. Despite this need, evaluating the performance of \Glspl{llm} in real-world scenarios remains a significant challenge. 
This is specially true in a conversational context, which is more complex than answering single-interaction requests.
Most current approaches to evaluate \Glspl{llm} focus primarily on specific tasks such as multi-QA \cite{zhuang2024toolqa, kamalloo2024towards} or 
code generation \cite{liu2024your, liu2024exploring}, 
which
do not fully evaluate the broader set of 
capabilities that \Glspl{llm} are expected to possess to truly function as an effective conversational 
AI agents.

Focusing on customer support, an effective AI agent is should be capable of 
interacting with tools and the customer in order to resolve
customer issues, while stricly adhering to procedures described by customer support admins. 
In order to assess the AI agent's performance, it is crucial to measure its ability to follow a given set of procedures and their resilience against potential customer manipulations. 
{For that,}
it is key to have a comprehensive evaluation dataset, which can lead to valuable insights into the agent's abilities and limitations.

\begin{figure*}
    \centering
    \includegraphics[width=\textwidth]{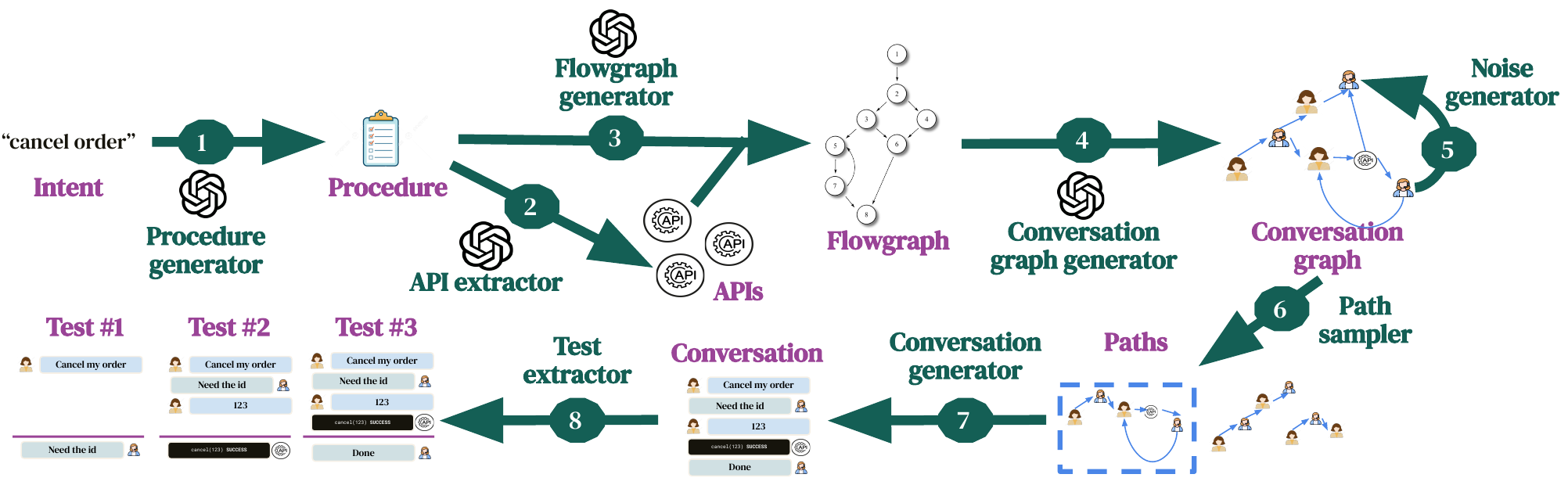}
        \caption{Automated test generation pipeline. For a given intent (e.g., cancel order) (1) we use an \GLS{llm} to generate a corresponding procedure. Then, (2) an \GLS{llm} extracts relevant APIs from the procedure, and (3) generates a flowgraph from the procedure and its APIs. Next, (4) an \GLS{llm} generates a conversation graph from the flowgraph and (5) adds noise to the graph (e.g., users going out of the expected procedure), to make the graph more realistic. To obtain conversations from the graph, (6) we sample paths from it, which correspond to different interactions. Finally, (7) an \GLS{llm} generates conversations from the paths and (8) we extract tests from the sampled conversations.} 
    \label{fig:cartoon}
\end{figure*}

We propose a method to generate evaluation datasets for tool-augmented LLMs as conversational AI agents. Our method automates dataset generation using an LLM to create conversations based on procedures, which are then transformed into tests. We use intermediate graph structures to improve the quality of the generated dataset (i.e., tests follow user-defined procedures) and make it more comprehensive (i.e., tests cover most relevant cases). To assess the AI agent's ability to handle attacks, we incorporate red teaming in our examples.

Our generation pipeline, illustrated in \Cref{fig:cartoon}, builds diverse datasets autonomously by using synthetically generated intents as seeds for procedures. Additionally, our pipeline also allows for the inclusion of real data where available, such as actual procedures or APIs used by a company to generate synthetic conversations. While datasets can be created fully automatically, we also put forward \texttt{ALMITA} (\textbf{A}utomated benchmark of \textbf{L}anguage \textbf{M}odels for \textbf{I}ntelligent \textbf{T}ool-augmented \textbf{A}gents), a manually curated dataset. We use this high-quality dataset to benchmark LLMs as conversational tool-augmented AI agents.

Our main contributions are: 



\begin{itemize}
    \item A method that generates datasets to evaluate tool-augmented LLMs as AI conversational agents, reducing manual effort needed to obtain such datasets. Our method provides an holistic evaluation of AI agents, with realistic and diverse conversations,  use of tools (e.g., functions/APIs), and grounded on user-defined procedures.
    \item \texttt{ALMITA}, the first conversational dataset that can be used to evaluate customer support AI agents, including both tooling (i.e., functions) and conversation reply to follow company user-defined procedures. \texttt{ALMITA} contains 1420 synthetic tests that were manually curated to ensure high-quality samples\footnote{\texttt{ALMITA}, along with all other datasets generated using our pipeline and referenced in the paper, are available in \url{https://github.com/zendesk/almita-dataset}.}. 
    \item Benchmarking of multiple LLMs on the proposed dataset. Our results indicate that current LLMs have high performance regarding single message accuracy and in calling the correct functions, but have limited accuracy when the complete conversation is considered, which might indicate that they would not be successful if deployed as fully autonomous customer support AI agents.
\end{itemize}


We also note that, while our evaluation focuses on customer support, the same method could be applied, with some changes, to other domains. 





\section{Related work}






With the increasing use of \Glspl{llm} as AI agents, significant efforts have been made to develop benchmarks to evaluate their ability to correctly answer customer requests in conversational settings. GAIA proposes 466 human-annotated questions covering tasks like general knowledge, daily tasks, and data analysis~\cite{Mialon2023GAIAAB}. Recently, AgentInstruct introduced a framework for generating synthetic data from diverse sources, such as code, web articles, and textbook chapters, to help agents generate and refine instruction sets~\cite{Mitra2024AgentInstructTG}. 
Unlike our work, these datasets do not assess tool-augmented AI agents.

Datasets to evaluate tool-augmented LLMs have been proposed. \citet{Zeng2023AgentTuningEG} propose AgentTuning and compile multiple agent datasets to create sequences of API calls. AgentBench features multi-step interactions between an agent and the environment, using various tools to solve user requests \cite{Liu2023AgentBenchEL}. \citet{patil2023gorilla} and \citet{qin2023toolllm} build datasets of APIs from sources like TorchHub, TensorHub, and rapidAI, prompting an LLM to generate instructions solvable by these APIs. 
\citet{basu2024api} combine multiple datasets to convert user instructions into API calls. APIGen introduced an automatic method to generate synthetic datasets for tool function calling \cite{liu2024apigen}. Unlike our work, these datasets are not conversational and just focus on mapping  utterances to API calls, and they do not use intermediate structures (i.e., graphs) to ensure coverage and reduce hallucinations in generated tests.

Other relevant work focuses on graph learning and on using different intermediate structures to reducing hallucinations. \citet{ye2023natural} propose InstructGLM, which uses natural language to describe node features used to tune an LLM for inference on graphs. \citet{wang2024can} introduce NLGraph, a benchmark for graph-based problems written in natural language, demonstrating that LLMs can perform structured operations on textual descriptions of graphs. Additionally, \citet{narayan2023conditional} propose using question-answer blueprints as intermediate representations to reduce hallucinations. These works do not fully encompass our problem setting of generating conversations in dialog format, calling APIs, and extracting  tests.



\begin{figure*}
    \centering
    \includegraphics[width=0.85\linewidth]{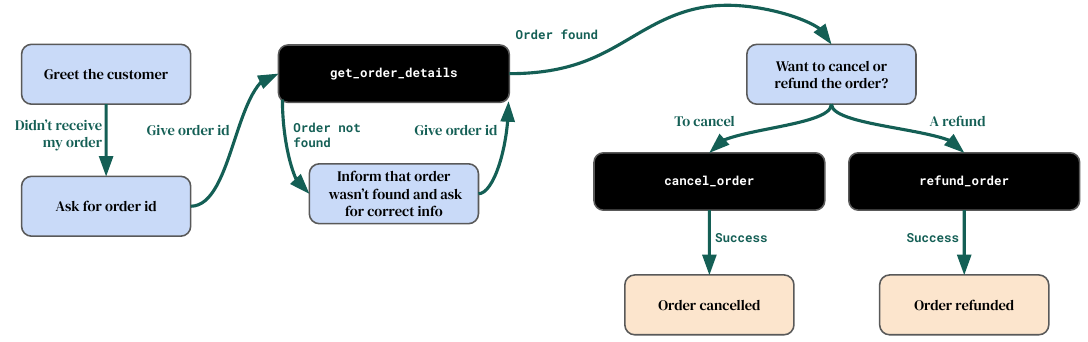}
    \caption{Flowgraph for intent \emph{Order not received} and procedure \emph{"If the customer did not receive their order, allow the customer to cancel or refund their order given that they provide a correct order id"}. 
    Blue nodes are message nodes, black nodes are API call nodes, orange nodes are end nodes. Edge labels are user messages or API outputs.} 
    \label{fig:flowgraph}
\end{figure*}

\section{Method}

Our automated test generation pipeline, illustrated in \Cref{fig:cartoon}, begins by generating textual procedures from input intents. While one could use an LLM to directly generate conversations from procedures, our approach converts the procedures into a flowgraph and then into a conversation graph. Our assumption is that using these intermediate structured representations makes the task of creating the conversations grounded on the procedures more accurate; see Section~\ref{sec:gen_from_conv} for supporting evidence. Additionally, the graphs allow us to introduce noise into the conversations, making conversations more realistic and challenging, and enable us to sample paths, ensuring path coverage and conversation diversity. We then generate conversations from the sampled paths. Finally, we extract tests from these conversations by breaking down the conversation at each user message, storing the context, and recording the generated response as the correct reply.

\subsection{Intent generator}
Intents (or \emph{issues}, e.g., cancel order) serve as the seeds for our automated test generation method. Intents can be generated by an LLM (as is the case in this work), sourced from predefined domain-specific intents, or a mix of both. The prompt used to generate intents is shown in \Cref{sec:intent_gen_prompt}.

\subsection{Procedure generator}\label{sec:proc_gen}

A procedure describes how a given issue/intent should be solved by an agent. We use an \GLS{llm} to generate a procedure for each input intent by asking it to provide a list of instructions that helps an agent fulfill a given task. We enforce in the prompt to avoid outputting general statements (e.g., "cancelling policies might depend on the company" or "explain the company's policy") since our goal is to generate specific and unambiguous procedures with precise and granular steps. We also enforce that conditionals are possible but that they need to have a clear solution in the  steps of the procedure. Finally, steps might contain actions based on APIs (e.g., search a database, escalate an issue) but they cannot be browsing actions (e.g., click on the login page). The full prompt is shown in \Cref{sec:procedure_gen_prompt}. 
Similarly to what we described for intents, existing procedures (e.g., of a company) can be included as input for our method. Moreover, procedures can be generated based on existing knowledge, namely existing tickets or help center articles.

{ Consider the intent \textit{"order not received"}: a simple procedure could be \textit{"If the customer did not receive their order, allow
the customer to cancel or refund their order given that they provide a correct order id"}. We use this procedure as an illustrative example throughout the paper (see  \Cref{fig:flowgraph,fig:conversation_graph,fig:test-extractor}.})

\subsection{API extractor}\label{sec:api_extractor}

Our target use-case is tool-augmented AI agents. We use an \GLS{llm} to generate APIs that are useful for an input procedure. We enforce in the prompt that the extracted APIs are agent APIs and not customer facing APIs. Generated APIs include not only the API name, but also their input output parameters, as well as a small description.
The full prompt is shown in \Cref{sec:api_gen_prompt}. These APIs should be explicitly called by the agent to fulfill the procedure. Similarly to intents and procedures, existing APIs can be easily included in our pipeline.

\subsection{Flowgraph generator}
\label{ssec:flowgraph-generator}

The flowgraph generator receives as input a procedure and relevant APIs and generates a directed graph encapsulating the logic of the procedure from the agent's perspective: nodes are agent actions and edges are customer replies or API outputs. Nodes are of 4 different types: (i) a single \texttt{start\_message} node is the initial message sent by the agent to the customer, (ii) \texttt{message} nodes are additional messages sent by the agent to the customer, (iii) \texttt{api} nodes are API calls performed by the agent, and (iv) \texttt{end\_message} nodes are messages by the agent that end the interaction. To reduce hallucinations and increase completeness, we enforce in the prompt (\Cref{sec:flowgraph_gen_prompt}) that every detail from the procedure needs to be in \texttt{message} nodes.

An example of a flowgraph is given in \Cref{fig:flowgraph}. Nodes in the flowgraph have a \texttt{node\_id} (e.g., "N1"), a \texttt{node\_type} (one of the four described above), and a \texttt{node\_description}, which should be related to a step in the procedure (e.g., "Tell the user the order was not found") or an API call (e.g., "\texttt{refund\_order}"). Edges in the graph are either the user interaction (e.g., "Gives order id and email") or the result of an API call (e.g., "Found order"). Edges in the flowgraph have an \texttt{edge\_id} (e.g., "E1"), a tuple with the source node and the target node (e.g., "(N1, N2)"), and an edge description, as described previously. We do one-shot prompting, providing an example to the \GLS{llm}; thus, a complete flowgraph can be seen in flowgraph prompt in \Cref{sec:flowgraph_gen_prompt}. 

To try to guarantee correct flowgraphs, we instruct the \GLS{llm} to generate graphs with only one root node with type \texttt{start\_message}, to always have concrete messages in the node and edge descriptions, and to provide API outputs in the outgoing edges of \texttt{api} nodes. To try to limit hallucinations and ensure that the graph encapsulates the entire procedure, we instruct the \GLS{llm} to follow strictly what is in the procedure and to include all content from it. At the end of the generation step, we convert the graph into a \texttt{networkx} graph and, if parsing succeeds, we pragmatically verify if all the rules described previously are followed; if they are not followed, we discard the generated flowgraph.

\subsection{Conversation graph generator}
\label{ssec:conversation-graph}

A flowgraph represents a sequence of agent steps to fulfill a procedure. The flowgraph's structure does not directly map to a conversation, which can make the task of creating a conversation from a flowgraph hard. Thus, the goal of the conversation graph generator is to convert the flowgraph into a a conversation graph, which is a structure that is more akin to a dialogue. The generated conversation graph is a directed graph that is expected to have nodes of three different types: (i) \texttt{agent} nodes are messages sent by the agent, (ii) \texttt{customer} nodes are messages sent by the customer, and (iii) \texttt{api} nodes are API calls by the agent. 

\begin{figure}
    \centering
    \includegraphics[width=\columnwidth]{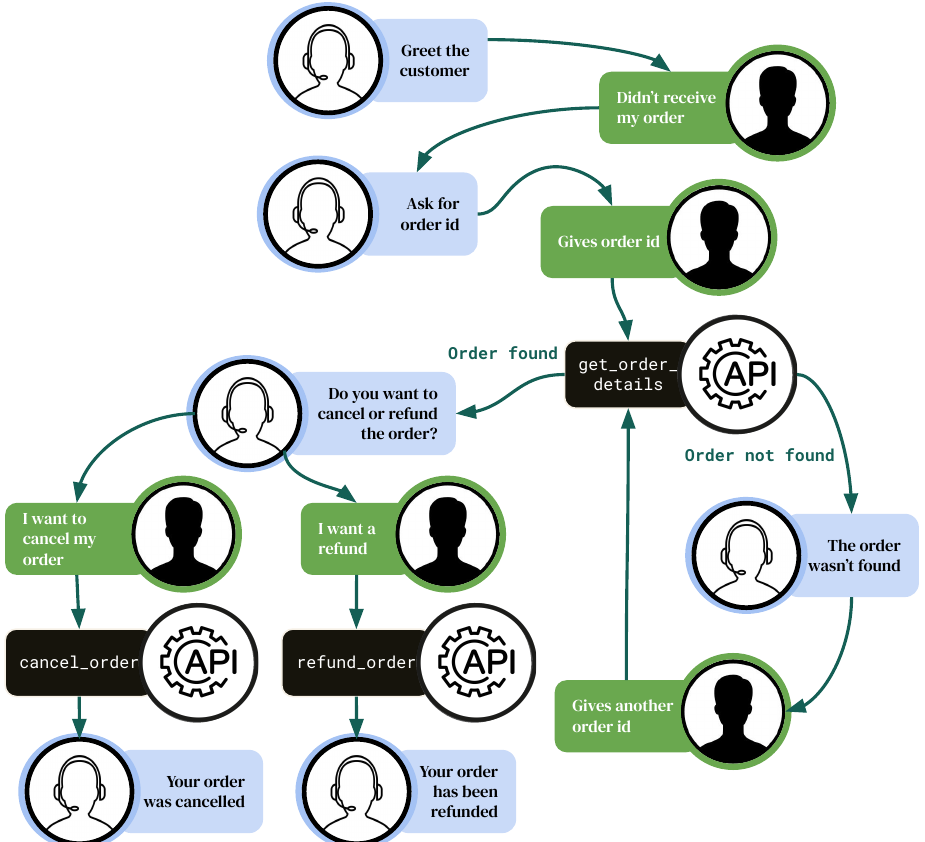}
    \caption{Conversation graph for flowgraph from Fig. \ref{fig:flowgraph} for intent \emph{Order not received}. Blue nodes are agent nodes, green are user nodes, and black are API nodes.} 
    \label{fig:conversation_graph}
\end{figure}

An example of a conversation graph is given in \Cref{fig:conversation_graph}.
Nodes in the conversation graph have a \texttt{node\_id} (e.g., "N1"), a \texttt{node\_type} (one of the three described above), and a \texttt{node\_description}, which is a message for \texttt{agent} and \texttt{customer} nodes, and an API call for \texttt{api} nodes. Edges in the conversation graph connect consecutive messages/api calls. Some conversation paths have conditions, such as an API call returning that the order was found or not; in these cases, edges have an edge description, otherwise the edge description is empty. Edges in the flowgraph have an \texttt{edge\_id} (e.g., "E1"), a tuple with the source node and the target node (e.g., "(N1, N2)"), and an edge description. 

In an effort to mitigate incorrect conversation graphs, we provide the \GLS{llm} with additional graph construction rules, e.g., \texttt{customer} nodes should be followed by either \texttt{agent} or \texttt{api} nodes, leaf nodes should be \texttt{assistant} nodes. We use one-shot prompting by giving the \GLS{llm} as input an example of a flowgraph and the corresponding conversation graph, as shown in \Cref{sec:con_graph_gen_prompt}. Similarly to flowgraphs, we load the generated graph into \texttt{networkx} and verify if the required conditions are met, otherwise the graph is discarded.

\subsection{Noise generator}

Conversation graphs are built from agent procedures, thus they are expected to only contain \emph{good} behaviour by both the agent and the customer (i.e., happy paths). To make AI agents more resilient to unexpected customer behaviour, which might be malicious or not, we augment the conversation graphs with behaviour outside of the procedure. 

The noise generator traverses the \texttt{agent} nodes in the conversation graph and, with a certain probability (e.g., 20\%), inserts an edge to a new \texttt{customer} node with a \texttt{node\_description} message which can either be an "out-of-procedure" message or an "attack" message. These messages are generated beforehand by an \GLS{llm}. Additionally, we add an edge from the noisy \texttt{customer} node to a new \texttt{agent} node with \texttt{node\_description} as "Say you're only here to help with the original issue." 

\begin{figure}[t]
    \centering
    \includegraphics[width=\columnwidth]{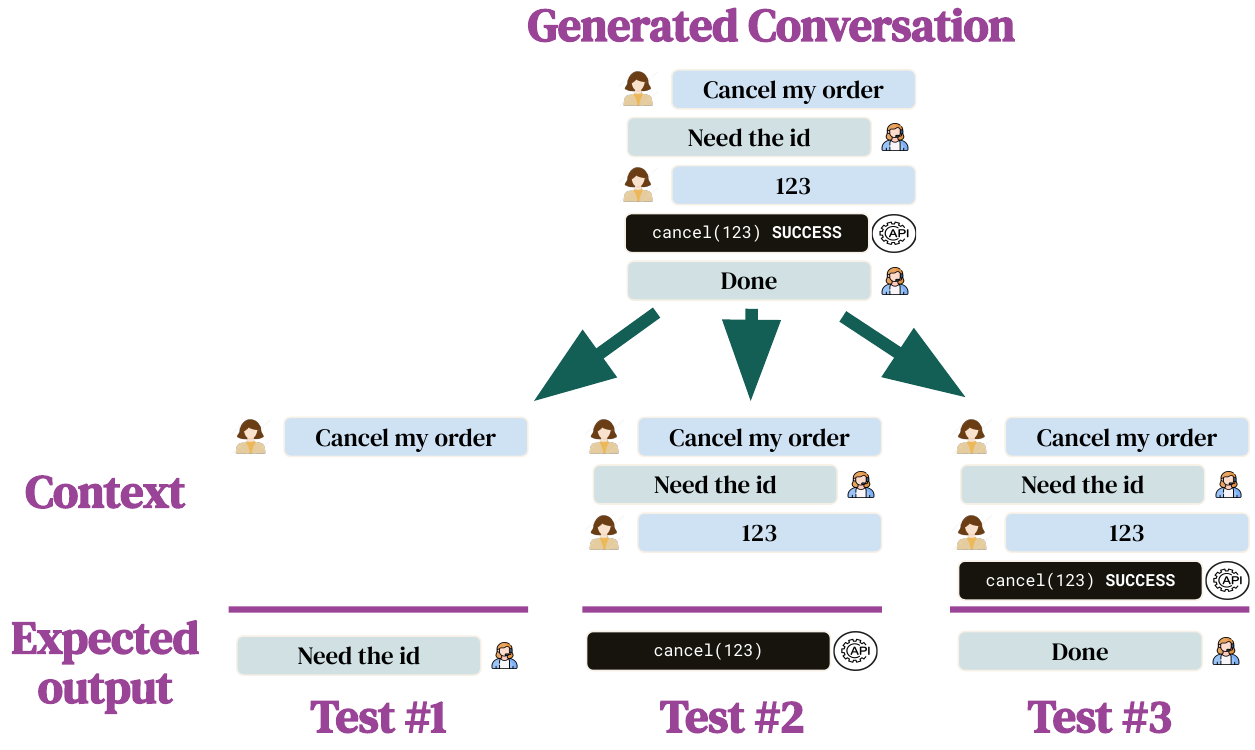}
    \caption{Tests extracted from one conversation.} 
    \label{fig:test-extractor}
\end{figure}

\subsection{Path sampler}\label{sec:path_sampler}

We extract conversations between a customer and an agent by sampling paths from the conversation graph. Given a conversation graph $\mathcal{G}$ with $N$ nodes and a desired number of conversations $M$, we employ a weighted random walks algorithm to sample paths, \Cref{alg:path_sampling}, which is an enhanced version of vanilla random walks, designed to improve node coverage. For that, we use a weighting vector $\bf{w}$ with $N$ elements initialized with ones (line 3). 
Each path $p$ is built by iteratively sampling nodes using $sample\_node$ (line 7). A node $n$, which is a child of the last node in the current path $p$, is sampled with a probability inversely proportional to its weight $w_i$, where $w_i$ is the number of times node $n$ was visited plus one (line 9). The index $i$ of node $n$ in graph $\mathcal{G}$ is provided by $node\_index$ (line 8). Path construction terminates when a leaf node is reached (lines 11--13).




\begin{algorithm}
\caption{Conversation path sampling}\label{alg:path_sampling}

\begin{algorithmic}[1]
\State \textbf{Inputs:} $\mathcal{G}$, $M$
\State $\mathcal{P} \gets \emptyset$
\State $\bf{w} \gets \bf{1}$$_N$

\State \textbf{while} $\lvert \mathcal{P} \rvert < M$ \textbf{do}
    \State \hspace{0.4cm} $p \gets \emptyset$
    
    \State \hspace{0.4cm} \textbf{while} True \textbf{do}
        \State \hspace{0.8cm} $n \gets \emph{sample\_node}(\mathcal{G}, p, \bf{w})$
        \State \hspace{0.8cm} $i \gets \emph{node\_index}(\mathcal{G},n)$
        \State \hspace{0.8cm} ${w_i} \gets {w_i} + 1$
        \State \hspace{0.8cm} $p \gets p \mid n$

        \State \hspace{0.8cm} \textbf{if} $n$ is EndNode \textbf{then}
            \State \hspace{1.2cm} $ \mathcal{P} \gets \mathcal{P} \mid p$
            \State \hspace{1.2cm} \textbf{break}
\end{algorithmic}
\end{algorithm}

\subsection{Conversation generator}\label{sec:conv_generator}

The conversation generator creates synthetic conversations from an input conversation graph, a sampled path, and relevant APIs. We provide the \GLS{llm} with context about the conversation graph structure and the APIs. Using one-shot prompting, we present the \GLS{llm} with an example triplet consisting of a conversation graph, a list of APIs, and a sampled path, as well as a possible conversation based on these conditions (see \Cref{sec:conversations_gen_prompt}). In an effort to generate valid conversations, we include conditions in the prompt, such as always generating a message with the API output following an API message, alternating customer and assistant messages, ensuring agents act on API output messages, and verifying API input and output types.


\begin{table*}[t]
    \centering
    \begin{tabular}{l|c|c|c|c|c|c|c}
         &Intents& Proc. & Proc. & Flowgraphs & Conv.
          Graphs & Conversations & Tests \\ 
         & & & w/ APIs & & & \\ \hline
         Generated 
         & 84 
         & 168 & 132 & 70 & 49 & 217 & 1,420 \\
         + auto. filters 
          & --
          & -- & 98 & 55 & 33 & -- & -- \\
         + man. 
         filters 
          & -- 
          & 132 & 70 & 49 & 33 & 192 & -- \\
          \texttt{ALMITA} & \bf 14 &\bf 18 &\bf 18 & \bf 18 & \bf 18 & \bf 192 & \bf 1,420 
         \\ \hline
        \texttt{auto-ALMITA} & \bf 52 & \bf63 & \bf63 & \bf63 & \bf63 & \bf407 & \bf2,696 
    \end{tabular}
    \caption{Statistics  while bootstrapping \texttt{ALMITA}'s dataset { from 84 intents}. We show the number of samples after (i) generation, (ii) automatic filtering, and (iii) human filtering annotations.
    "--" indicates no filtering. \texttt{auto-ALMITA} was created using the same 84 seed intents as \texttt{ALMITA}, but using the same pipeline without any human filtering, so that we can assess the capabilities of our test generation
pipeline when no human annotators are available.
    }
    \label{tab:dataset_statistics}
\end{table*}
\subsection{Test extractor}\label{sec:test_extractor}

The test extractor converts a single conversation into one or more tests. It iteratively breaks down the conversation into sub-conversations (or contexts), each ending with a customer message (e.g., "Cancel my order") or an API output (e.g., "success" following a cancel function call). The rationale is that since the generated conversations exemplify correct flows, we can construct contexts using the preceding messages, with the expected output being the next non-customer message, whether it’s an agent response or an API call. \Cref{fig:test-extractor} illustrates an example of three tests extracted from a generated conversation. Tests are used to evaluate an AI agent by providing it with the context and comparing its response with the expected output.

\section{Results}

\input{sections/results}

\section{Conclusions}

LLMs are being used as customer support AI agents. However, existing evaluation datasets are limited in their scope. We propose an automated test generation pipeline to evaluate tool-augmented conversational AI agents. Our proposed method uses LLMs to generate intermediate graph structures that help limit hallucinations and improve diversity in the generated tests. We evaluate different LLMs to analyze the current capabilities of LLMs implemented as AI agents.

To facilitate this, we developed the \texttt{ALMITA} dataset, which we used to thoroughly evaluate these AI agents and identify their limitations. \texttt{ALMITA} allows for a multifaceted evaluation across several key dimensions, such as reply accuracy, API call correctness, and overall conversation integrity. Our findings highlighted significant limitations in current LLMs, particularly in maintaining correct conversations throughout a user interaction.

Importantly, the \texttt{ALMITA} dataset can be used by other researchers to evaluate AI agents, providing a comprehensive benchmark for assessing various aspects of their performance, possibly in other target domains. Additionally, since our test generation pipeline is fully automated, we have the capability to create new, more challenging versions of the dataset. This adaptability ensures that our framework can be continually updated to reflect more complex and realistic scenarios, further enhancing its utility for ongoing research and development of AI agents in customer support and beyond.


\section{Limitations}

Our evaluation has some limitations. Namely, we did not evaluate the diversity of the generated tests quantitatively. We performed human annotation, to verify correctness at each step, but the number of annotations and of annotators was small. Our test generation pipeline only used a single LLM as the generator, namely GPT4
and this might influence evaluation. A possible mitigation for this is to repeat the test generation pipeline for other LLMs and aggregate the tests. We evaluated multiple LLMs but only using a single prompt. Our goal was to test different models on the generated dataset, but more advanced AI agents could be considered.

Additionally, we acknowledge that some metrics may be too strict.  
As a future direction, we would like to consider the severity of the errors of an AI agent in a conversation. Conversations are relatively fluid and we may have other replies/actions that are somehow acceptable for a given procedure besides of the most obvious and direct one that was annotated in the dataset. There is still to be develop more advanced and more semantic conversational metrics allowing for some path variations, similarly to what has been happening for the comparison of two sentences where different words and order of words can lead to similar meanings.

\bibliography{anthology,custom}
\bibliographystyle{acl_natbib}

\clearpage

\appendix
\label{sec:appendix}

\section{Prompts}
\subsection{Intent generation}\label{sec:intent_gen_prompt}
\input{prompts/issue_system_prompt}
\input{prompts/issue_user_prompt}

\subsection{Procedure generation}\label{sec:procedure_gen_prompt}
\input{prompts/procedure_generation_system_prompt}
\input{prompts/procedure_generation_user_prompt}

\subsection{API extraction}\label{sec:api_gen_prompt}
\input{prompts/api_extraction_system_prompt}
\input{prompts/api_extraction_user_prompt}

\subsection{Flowgraph generation}\label{sec:flowgraph_gen_prompt}
\input{prompts/flowgraph_generator_system_prompt}
\input{prompts/flowgraph_generator_user_prompt}

\subsection{Conversation graph generation}\label{sec:con_graph_gen_prompt}
\input{prompts/conversation_graph_generator_system_prompt}
\input{prompts/conversation_graph_generator_user_prompt}

\subsection{Conversations generation}\label{sec:conversations_gen_prompt}
\input{prompts/conversations_generator_system_prompt}
\input{prompts/conversations_generator_user_prompt}

\subsection{Conversations from procedures}\label{sec:conversations_straight_through_gen_prompt}
\input{prompts/conversation_straight_through_generation_system_prompt}
\input{prompts/conversation_straight_through_generation_user_prompt}

\subsection{Tool-augmented AI agent}\label{sec:agent_prompt}
\input{prompts/agent_system_prompt}
\input{prompts/agent_user_prompt}


\section{\texttt{auto-ALMITA}: Detailed evaluation}

Supplementary Table~\ref{tab:agents_auto-eval} provides detailed results obtained with the \texttt{auto-ALMITA} dataset, considering the 6 LLM agents and all the evaluation metrics from \Cref{sec:agent_evaluation}.

\captionsetup[table]{name=Supplementary Table}
\setcounter{table}{0}

\begin{table*}[h]
    \centering
    \begin{tabular}{l|c|c|c|c|c|c|c}
         \multirow{2}{*}{LLM}    & \multicolumn{2}{c|}{Reply}  & \multicolumn{3}{c|}{API}           & Test    & Conversation  \\
                                 & Recall     & Correct        & Recall    & Correct  & Correct params. & Correct  & Correct  \\
         \hline
         \texttt{GPT-4o}         & 91.1       & 77.1           & 89.5      & 95.1     & 84,4            & \bf 85.4 & \bf 14.7 \\
         \texttt{Mistral-NeMo-I} & 89.2       & 67.5           & 89.5      & 93.8     & 80.7            & 81.3     & 10.3     \\
         \texttt{Claude3-s}      & 79.9       & 67.1           & 92.9      & \bf 95.9 & 84.1            & 78.9     & 6.9      \\
         \texttt{GPT-4}          & 60.5       & \bf 82.9       & 92.6      & 94.6     & \bf 84.5        & 75.5     & 6.4      \\
         \texttt{Llama3.1-8b-I}  & 79.4       & 61.8           & 64.3      & 95.7     & 83.8            & 73.4     & 3.2      \\
         \hline
         \texttt{GPT-4o w/ F}    & \bf 89.6   & 75.3           & \bf 93.0  & 93.8     & 72.2            & 82.9     & 11.5     \\
    \end{tabular}
    \caption{LLM AI agents evaluated on \texttt{auto-ALMITA}.  For each LLM, the highest value in shown in \textbf{bold}. All results are percentages.}
    \label{tab:agents_auto-eval}
\end{table*}

\end{document}

%% file: acronyms.tex
\newacronym{llm}{LLM}{large language model}

%% file: sections/results.tex
\begin{table*}
    \centering
    \begin{tabular}{l|c|c|c|c|c|c|c}
          \multirow{2}{*}{LLM} & \multicolumn{2}{c|}{Reply}  & \multicolumn{3}{c|}{API}   & Test & Conversation \\
           & Recall & Correct & Recall & Correct & Correct params. & Correct & Correct \\ \hline
         \texttt{GPT-4o} & 92.7 & 75.2 & 96.7 & \bf 99.8 & 92.2 & \bf 88.9 & 14.1 \\
         \texttt{Mistral-NeMo-I} & 92.0 & 65.0 & 89.8 & 99.5 & 92.1 & 84.7 & 7.3 \\
         \texttt{Claude3-s} & 88.0 & 60.3 & 96.2 & \bf 99.8 & 90.5 & 83.3 & 10.4 \\
         \texttt{GPT-4} & 53.2 & \bf 77.7 & \bf 98.1 & \bf 99.8 & \bf 93.0 & 76.9 & 4.2 \\
         \texttt{Llama3.1-8b-I} & 74.8 & 53.5 & 72.1 & 90.8 & 85.9 & 73.1 & 1.6 \\ \hline
         \texttt{GPT-4o w/ F} & \bf 92.9 & 74.8 & 97.2 & 99.0 & 86.6 & 88.0 & \bf 15.6 \\
    \end{tabular}
    \caption{AI agents evaluated on their capacity to produce correct replies with correct API calls. We test different LLMs using the same prompt. Additionally, we evaluate LLMs using function calling (with the "\texttt{w/ F}" suffix). The versions of the closed source models are \textit{gpt-4-0613}, \textit{gpt-4o-2024-05-13}, \textit{anthropic.claude-3-sonnet-20240229-v1:0}. The "\texttt{-I}" suffix indicates that it is an instruction model. All results are percentages, with the highest value in \textbf{bold}.}
    \label{tab:agents_eval}
\end{table*}

In Section~\ref{sec:gen_dataset} we detail the creation of \texttt{ALMITA}, a manually curated dataset for evaluating LLMs as AI customer support agents. Two annotators independently review each datapoint to identify incorrect instances, followed by a discussion to align their assessments and minimize disagreements. Any datapoint deemed incorrect by at least one annotator is then removed.
GPT-4 is used for all generation steps (see Figure~\ref{fig:cartoon}). 
To assess the benefits of the graph intermediate structures, we conduct an ablation study comparing conversations generated directly from procedures to those using the intermediate structures, with manual curation for quality assessment (Section~\ref{sec:gen_from_conv}). 
In Section~\ref{sec:agent_evaluation}, we evaluate various AI agents on \texttt{ALMITA}.
Finally, in Section~\ref{sec:auto_almita}, we assess the effectiveness of our pipeline in generating high-quality test sets automatically. We do this by comparing the AI agents' performance on \texttt{ALMITA} with those on its fully automated counterpart, \texttt{auto-ALMITA}.



\subsection{Dataset generation: \texttt{ALMITA}}\label{sec:gen_dataset}

We begin by asking the LLM to generate intents using the prompt from \Cref{sec:intent_gen_prompt}, resulting in 84 intents. Using them as input, we prompt the model to generate two procedures per intent, for a total of 168 procedures. After manual annotation, we remove 36 procedures that did not comply with the rules from \Cref{sec:proc_gen}. The valid procedures average 315 words (ranging from 171 to 535) and 11 steps (ranging from 6 to 19). Next, we extract APIs for each procedure as outlined in \Cref{sec:api_extractor}. APIs not in the correct JSON format are automatically filtered out, along with procedures with invalid APIs, resulting in 70 valid procedures. Each of these procedures, on average, includes 4 APIs (ranging from 2 to 9). For each of the 70 procedures with APIs, we generate the corresponding flowgraph. We automatically filter out 15 flowgraphs and manually filter 6 more that do not adhere to the rules discussed in \Cref{ssec:flowgraph-generator}. The valid flowgraphs average 15 nodes (ranging from 10 to 20) and 17 edges (ranging from 10 to 25). For each of the remaining 49 valid flowgraphs, we generate the corresponding conversation graph. We automatically exclude 16 conversation graphs and manually exclude 7 more based on adherence to rules (\Cref{ssec:conversation-graph}). The valid conversation graphs average 23 nodes (ranging from 16 to 37) and 24 edges (ranging from 15 to 37). From these conversation graphs, we generate 217 conversations after path sampling (\Cref{sec:path_sampler}). We manually filter out 25 conversations for not following the rules (\Cref{sec:conv_generator}). Thus, from the original 84 intents, we obtain 192 valid conversations. Each conversation traverses an average of 12 nodes (ranging from 3 to 24). Finally, tests are extracted from these conversations as detailed in Section~\ref{sec:test_extractor}, resulting in 1420 generated tests. Table~\ref{tab:dataset_statistics} summarizes the dataset statistics. In the end, the \texttt{ALMITA} dataset comprises 14 intents, 18 procedures, 18 flowgraphs, 18 conversations graphs, 192 conversations and 1420 tests.




\subsection{Ablation study: conversations from procedures}\label{sec:gen_from_conv}

We conduct an ablation study to validate the effectiveness of our intermediate graph representations in generating correct conversations. We remove the flowgraph generator, conversation graph generator, noise generator, and path sampler, and generate conversations directly from the procedures and APIs using the prompt from \Cref{sec:conversations_straight_through_gen_prompt}. 
Annotating conversations directly generated from procedures showed to be a much more complex and time-consuming than annotating conversations generated from graphs. For this reason we only annotate 50 conversations. 
{All 50 conversations are generated from the same 70 input procedures as \texttt{ALMITA}, and they are curated by the same two annotators, following the same annotation strategy}.
K
The simplified pipeline results in  
$\approx68\%$ ($34/50$)
valid conversations, as evaluated by the same annotators that curated \texttt{ALMITA}. In contrast, the original pipeline with intermediate graph representations yields 
$\approx88\%$ ($192/217$) valid conversations. This indicates 
that graph representations improve the validity of 
generated conversations.
Even when considering the cumulative impact of curating flowgraphs, the original pipeline would automatically generate 
$\approx78\%$ ($192/217 \times 49/55$) valid conversations, which is above  $\approx68\%$.


Moreover, while the prompt used in the simplified pipeline could potentially be improved, the simplified pipeline intrinsically does not ensure that all branching paths from the procedure are explored. This highlights the benefit of intermediate graph representations in covering all possible conversation paths.



\subsection{Evaluation of LLM AI agents}\label{sec:agent_evaluation}

We use \texttt{ALMITA} to evaluate LLMs serving as customer support AI agents. The dataset allows us to evaluate the following dimensions, which we report in \Cref{tab:agents_eval}: (i) \emph{reply recall}: when the correct action is to reply, the agent correctly sends a reply message instead of calling an unnecessary API, (ii) \emph{correct reply}: when both the correct and the predicted action is to reply, the agent's reply matches the expected reply (we use BERTScore with a similarity threshold of 0.55 after inspecting of some examples), (iii) \emph{API recall}: when the correct action is to do an API call, the agent correctly detects that it needed to perform an API call instead of replying, (iv) \emph{correct API}: when both the correct and the predicted action is to perform an API call, the agent calls the correct API; (v) \emph{correct API parameters}: when both the correct and the predicted action are the same API call, the agents calls the API with the correct parameter values, (vi) \emph{test correctness (or test accuracy)}: whether the test is fully correct (i.e., call the correct reply/API and, if the correct action is an API, call the correct API and use the correct parameters, or if the correct action is a reply, provide a correct reply), (vii) \emph{conversation correctness (or conversation accuracy)}: whether the sequence of all tests from the conversation where all correct. 

We evaluate 5 different LLMs: \texttt{GPT4-o}, \texttt{GPT-4}, \texttt{Claude3-sonnet}, \texttt{Mistral-NeMo-Instruct}, and \texttt{Llama3.1-8b-Instruct}. To ensure fairness, we use a uniform prompt for all models 
(details in \Cref{sec:agent_prompt}).
Our 
prompt aims to be general, avoiding any favoritism towards a specific model, although we acknowledge that different models may excel with different prompting styles. 
Since
the dataset 
includes 
API calling, we also test \texttt{GPT4-o} with function calling, 
denoted as 
\texttt{GPT-4o w/F}.


We observe that all LLMs demonstrate high accuracy when responding with an API, achieving over 85\% correctness in both the \emph{correct API} and \emph{correct API parameters} dimensions. With the exception of \texttt{Llama3.1-8b-I}, which performs considerably worse, the other models correctly determine when an API should be called, with an \emph{API recall} exceeding 90\%. However, performance in other dimensions is notably lower, suggesting that datasets focused solely on API calls do not comprehensively evaluate an AI agent's capabilities.

Interestingly, \texttt{GPT-4} tends to call APIs even when unnecessary, resulting in a lower \emph{reply recall} compared to other models. In terms of \emph{correct reply}, \texttt{GPT} models outperform the others, though this may be biased by the use of \texttt{GPT-4} for test generation. For \emph{test correctness}, \texttt{GPT-4o}, \texttt{Claude3-s}, and \texttt{Mistral-NeMo-Instruct} show the highest performance, while \texttt{GPT-4} and \texttt{Llama3.1-8b-Instruct} rank among the lowest.

Most critically, we see that all models have very low performance regarding \emph{correct conversation}. In practice, this would mean that these AI agents would very likely fail at some step of a conversation with a user. This showcases that current LLMs have some limitations that require either better models or very engineered prompts to suitably serve as fully autonomous customer support AI agents.

Our dataset could, potentially, be useful  to evaluate future models and/or strategies on their AI agent capabilities.
Furthermore, since the pipeline is automated, the dataset could be updated to include more (and harder) tests, as well as adapted to new or more specific domains.

\subsection{Fully automated tests: \texttt{auto-ALMITA} }\label{sec:auto_almita}


In this section, we analyze the results obtained by AI agents on \texttt{auto-ALMITA}, the fully automated version of the \texttt{ALMITA} dataset. This dataset was created using the same seed intents from the \texttt{ALMITA} dataset, described in  \cref{sec:gen_dataset}. Then we run the same pipeline without the manual filtering steps.
\texttt{Auto-ALMITA} retains more data points and greater diversity (see Table~\ref{tab:dataset_statistics}), albeit with some reduction in quality. Being fully automatically generated, \texttt{auto-ALMITA} can also be easily extended without additional curation efforts.

\begin{figure}
\pgfplotsset{compat=1.18}
\begin{tikzpicture}
    \begin{axis}[
        width=\linewidth,
        height=6cm, 
        xlabel={\emph{test correct} @ \texttt{ALMITA}},
        ylabel={\emph{test correct} @ \texttt{auto-ALMITA}},
        xmin=69, xmax=95,
        ymin=69, ymax=95,
        xtick={70, 74, 78, 82, 86, 90, 94},
        ytick={70, 74, 78, 82, 86, 90, 94},
        legend pos=north west,
        grid=major,
        grid style={line width=0.2pt, opacity=0.5},
        label style={font=\small}, 
        ticklabel style={font=\small}, 
        scatter/classes={
            a={mark=*, draw=black, fill=orange, opacity=0.0}, 
            b={mark=*,draw=black,fill=purple, opacity=0.0}, 
            c={mark=*,draw=black,fill=green, opacity=0.0}, 
            d={mark=*,draw=black,fill=red, opacity=0.0}, 
            e={mark=*,draw=black,fill=brown, opacity=0.0}, 
            f={mark=*,draw=black,fill=blue, opacity=0.0} 
        },
        tick style={draw=none}, 
        every tick label/.append style={font=\scriptsize, color=transparent}
    ]
    \addplot[scatter,only marks,scatter src=explicit symbolic] 
        coordinates {
        (77, 75.5) [a] 
        (73.5, 73.5) [b] 
        (89, 85.5) [c] 
        (88, 83) [d] 
        (84.5, 81.5) [e] 
        (83.5, 78.5) [f] 
    };
    
    \node at (axis cs:77, 75.5) [anchor=west, font=\scriptsize] {\hspace{-0.25cm}\includegraphics[width=0.2cm]{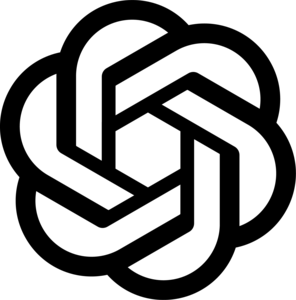} \texttt{GPT-4}};
    \node at (axis cs:73.5,73.5) [anchor=west, font=\scriptsize] {\hspace{-0.25cm} 
 \includegraphics[width=0.3cm]{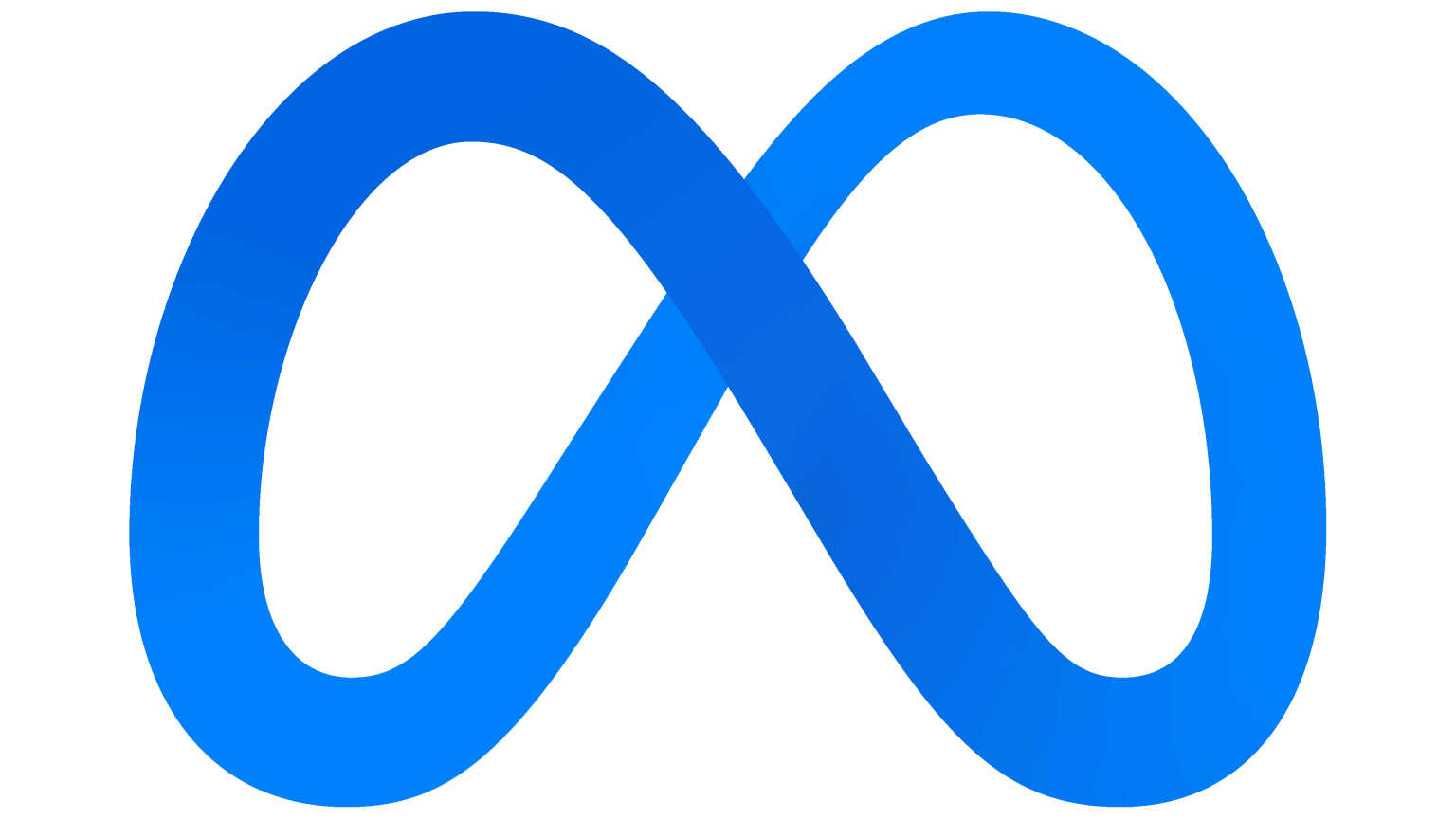} 
 \texttt{Llama3.1-8b-I}};
    \node at (axis cs:89, 85.5) [anchor=west, font=\scriptsize] {\hspace{-0.25cm}\includegraphics[width=0.2cm]{images/gpt4.png} \texttt{GPT-4o}};
    \node at (axis cs:88,83) [anchor=west, font=\scriptsize] {\hspace{-0.25cm} \includegraphics[width=0.2cm]{images/gpt4.png} \texttt{GPT-4o w/ F}};
    \node at (axis cs:84.5,81.5) [anchor=west, font=\scriptsize] {\hspace{-0.25cm} 
 \includegraphics[width=0.3cm]{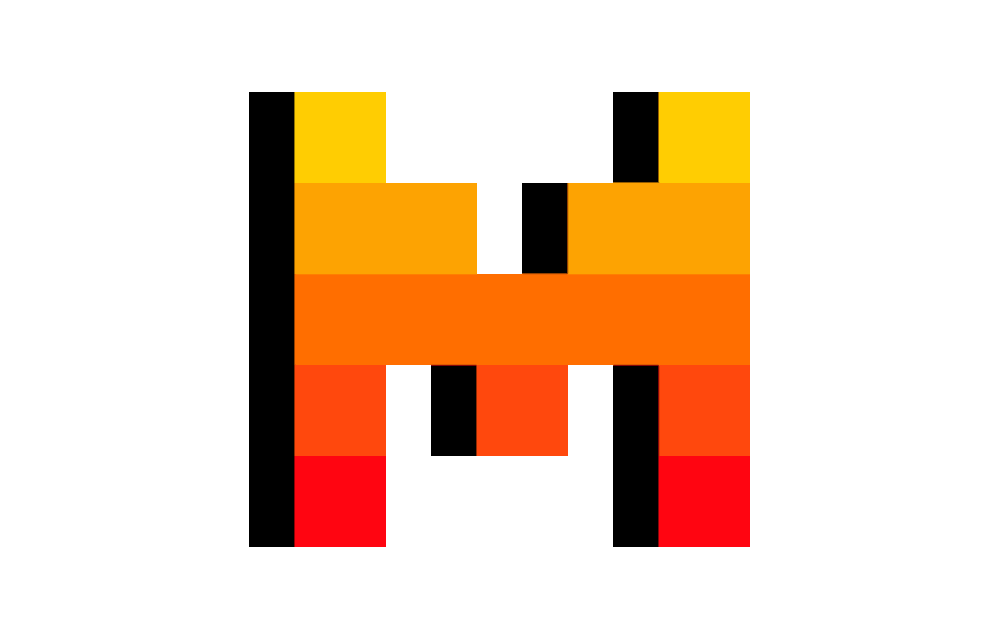}\texttt{Mistral-NeMo-I}};
    \node at (axis cs:83.5, 78.5) [anchor=west, font=\scriptsize] {\hspace{-0.25cm} 
 \includegraphics[width=0.2cm]{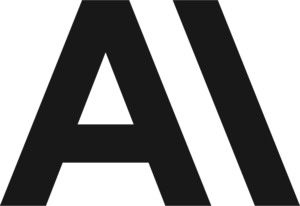} \texttt{Claude3-sonnet}};
    
    \end{axis}
\end{tikzpicture}
    \caption{
    \emph{test correct} value for different LLM Agents on the \texttt{auto-ALMITA} and \texttt{ALMITA} datasets. 
    } 
    \label{fig:almita_vs_auto_almita}
\end{figure}

We evaluate the same LLM agents from \Cref{tab:agents_eval} and compare the global metric \textit{test correct} obtained by the AI agents both \texttt{auto-ALMITA} and \texttt{ALMITA} in \Cref{fig:almita_vs_auto_almita}. Both datasets rank the LLMs in the same order, with a high correlation value of 0.98 (detailed results are provided in Supplementary Table~\ref{tab:agents_auto-eval}). These findings suggest that the proposed pipeline can generate evaluation datasets for AI agents entirely automatically, which lead to conclusions similar to those derived from curated datasets.

%% file: prompts/issue_system_prompt.tex
\begin{tcolorbox}[colback=black!5!white, colframe=black!75!black, title=System prompt]
\scriptsize 
You are <REDACTED>, a platform providing customer support. You serve clients from numerous different industries: internet providers, financial institutions, e-commerce platforms, entertainment websites, etc. All these clients have customer that can contact customer support to obtain information, complain about something, or other reasons to contact the customer support team.
\end{tcolorbox}

%% file: prompts/issue_user_prompt.tex
\begin{tcolorbox}[
    colback=black!5!white, 
    colframe=black!75!black, 
    title={User prompt},
]
\scriptsize 
Your task is to generate a list of problems that can lead to a customer contacting support. Think of the type of client for which the issue is relevant, a description of the detailed issue, and a short name for the error.

Generate \{\{ number\_issues \}\} issues from a diverse pool of clients.

Format your answer as a json with the following structure:
\begin{lstlisting}[basicstyle=\ttfamily\fontsize{6}{6}\selectfont]
[{
   "client": e.g., a bank, internet provider, etc. Do not limit yourself to these examples!,
   "issue": description of the error, be specific!,
   "name": a short name for the issue
}]
\end{lstlisting}
\end{tcolorbox}

%% file: prompts/procedure_generation_system_prompt.tex
\begin{tcolorbox}[colback=black!5!white, colframe=black!75!black, title=System prompt]
\scriptsize 
You are <REDACTED>, a platform providing customer support. You serve clients from numerous different industries: internet providers, financial institutions, e-commerce platforms, entertainment websites, etc. All these clients have customer that can contact customer support to obtain information, complain about something, or other reasons to contact the customer support team.

Your task is to generate a procedure that helps an agent to fulfil a task. The agent can take actions or they can ask the customer for data (e.g., email address). You can include branching in the procedure.

\vspace{0.2cm}
Do not give general statements such as "Each system might have different processes". Instead, assume the role of a specific company that has very defined processes.

Do not give general steps such as "Explain the company's policy". The agent is following a procedure, so all steps need to be clearly stated, e.g., state precisely what is the policy. Do not leave room for ambiguity nor lack of information.

Do not state conditionals that are not resolved in the procedures such as "If it is allowed by the policy". Every conditional has to be fully contained in the procedure, the agent should not have to read another document nor rely on other knowledge about the company's procedures. Your role is to make up reasonable scenarios that are unambiguous.

Steps should be precise and granular.

Avoid giving examples, we want a concise procedure.

Do not include actions that are unrelated to the interaction with the client (e.g., document the interaction, monitor the process). The procedure is solely on how to address the issue reported by the customer.

Assume that you don't have a browser. Do not include navigation steps, just the actions that the agent should take.
\end{tcolorbox}

%% file: prompts/procedure_generation_user_prompt.tex
\begin{tcolorbox}[
    colback=black!5!white, 
    colframe=black!75!black, 
    title={User prompt},
]
\scriptsize 
\begin{lstlisting}[basicstyle=\ttfamily\fontsize{6}{6}\selectfont]
# Issue
{{ issue }}
\end{lstlisting}
\end{tcolorbox}

%% file: prompts/api_extraction_system_prompt.tex
\begin{tcolorbox}[colback=black!5!white, colframe=black!75!black, title=System prompt]
\scriptsize
You are a programming assistant working for a customer experience company. Given a procedure an agent should follow to solve a customer problem, your job is to extract ALL possible APIs used by the agent.

Never generate an API call that asks for passwords.
The APIs should be as specific as possible to what is in the procedure and not general methods.
All the API parameters should have type different than None. When representing structured output follow python convention like list[str] or dict[str, float].
Optional parameters should follow the python convetion of Optional[str].
If the procedure doesn't have any action an agent should solve, return an empty JSON.

\begin{lstlisting}[basicstyle=\ttfamily\fontsize{6}{6}\selectfont]
# Output
Respond only in JSON format with the following schema. The name of the api should be written in snake case.
{"apis": [{"name": str, "desc": str, "params": [{"name": str}],  "output": {"name": str, "type": str}}]}
\end{lstlisting}
\end{tcolorbox}

%% file: prompts/api_extraction_user_prompt.tex
\begin{tcolorbox}[
    colback=black!5!white, 
    colframe=black!75!black, 
    title={User prompt},
]
\begin{lstlisting}[basicstyle=\ttfamily\fontsize{6}{6}\selectfont]
# Procedure
```
{{ procedure }}
```
\end{lstlisting}
\end{tcolorbox}

%% file: prompts/flowgraph_generator_system_prompt.tex
\begin{tcolorbox}[colback=black!5!white, colframe=black!75!black, title=System prompt]
\scriptsize 
You are and experienced flowchart creator. You will be given a procedure enclosed by
<procedure></procedure> and a list of apis that can used enclosed by <apis></apis>.
Your job is to extract the flowchart used to solve the problem. Your flowchart will
be used by an assistant to know how to solve the problem. The agent has no access to the procedure,
so all the information has to be contained in this flowchart!!

You are and experienced flowchart creator. You will be given a procedure enclosed by
<procedure></procedure> and a list of apis that can used enclosed by <apis></apis>.
Your job is to extract the flowchart used to solve the problem. Your flowchart will
be used by an assistant to know how to solve the problem. The agent has no access to the procedure,
so all the information has to be contained in this flowchart!!
\vspace{0.2cm}

The flowchart is constituted by nodes and edges in the following format:

\begin{lstlisting}[basicstyle=\ttfamily\fontsize{6}{6}\selectfont]
[node_id](node_type){node_description}
[edge_id](parent_node_id, child_node_id){edge_description}
\end{lstlisting}

Node ids should always be N followed by an integer. Edge ids should always be E followed by an integer.

You can use nodes of the type start\_message, message, api and end\_message.

- start\_message: initial message sent by the assistant to the customer, taken from the procedure. It doesn't have a parent node.

- message: node with a message sent by an assistant to the customer. this message should have all the details found in the procedure.

- api: api call the assistant should perform.

- end\_message: node to send a message and finish execution.

\vspace{0.2cm}
Graph construction rules

- The graph only have one root node of type `start\_message`.

- An outgoing edge from a message node is the reply of the customer. Customer messages have to be specific.

- An outgoing edge from an api node is the output of the api.

- End nodes cannot have outgoing edges and should be of type end\_message.

- End nodes have the node type `end\_message`.

- Never have an edge going back to the start node N0.

\end{tcolorbox}

\begin{tcolorbox}[colback=black!5!white, colframe=black!75!black]
\scriptsize 
Details

The messages by the agent and the customer should follow stricly what is in the procedure. ALL the details in the procedure need to be in the flowchat! Don't assume that the agent will ever see the procedure, so it is critical that the details are here, such as reasons for something to fail, or information that needs to be collected.

Make sure all steps are nodes. Some procedures might have branching paths.

Always use the APIs when appropriate.

The flowchart must be enclosed by <flow></flow>.

\begin{lstlisting}[basicstyle=\ttfamily\fontsize{6}{6}\selectfont]
Example of a flow:
<flow>
[N0](start_message){Greet the customer}
[E0](N0, N1){Didn't receive my order}
[N1](message){Ask customer for order id, the email or phone number}
[E2](N1, N2){Gives order id and email}
[E3](N1, N3){Gives order id and phone number}
[N2](api){get_order_details_by_email}
[N3](api){get_order_details_by_phone_number}
[N4](message){Do you want to cancel or refund the order?}
[E3](N2, N4){Found order}
[E5](N3, N4){Found order}
[N5](message){Tell the user the order wasn't found and ask for correct information}
[E5](N2, N5){Order not Found}
[E6](N3, N5){Order not Found}
[E6](N5, N2){User provides another email or order id}
[E7](N5, N3){User provides another phone number or order id}
[N6](api){cancel_order}
[E8](N4, N6){I want to cancel the order}
[N7](end_message){Order cancelled}
[E9](N6, N7){Success}
[N8](api){refund_order}
[E9](N4, N8){I want a refund}
[N9](end_message){Order refunded}
[E10](N8, N9){Success}
</flow>

<apis>
{{ apis }}
</apis>
\end{lstlisting}
\end{tcolorbox}

%% file: prompts/flowgraph_generator_user_prompt.tex
\begin{tcolorbox}[
    colback=black!5!white, 
    colframe=black!75!black, 
    title={User prompt},
]
\begin{lstlisting}[basicstyle=\ttfamily\fontsize{6}{6}\selectfont]
<procedure>
{{ procedure }}
</procedure>
\end{lstlisting}
\end{tcolorbox}

%% file: prompts/conversation_graph_generator_system_prompt.tex
\begin{tcolorbox}[colback=black!5!white, colframe=black!75!black, title=System prompt]
\scriptsize 
Your task is to convert a flowchart into a conversation graph. 

The flowchart will be given in between <flowchart></flowchart>.

The flowchart is constituted by nodes and edges in the following format:

\begin{lstlisting}[basicstyle=\ttfamily\fontsize{6}{6}\selectfont]
[node_id](node_type){node_description}
[edge_id](parent_node_id, child_node_id){edge_description}
\end{lstlisting}

Nodes are of the following types:

- start\_message: initial message sent by the assistant to the customer, taken from the procedure.

- message: node with a message sent by an assistant to the customer.

- api: api call the assistant should perform.

- end\_message: node to send an assistant message and finish execution.
\vspace{0.2cm}
You need to convert it into a conversation graph where:

\begin{lstlisting}[basicstyle=\ttfamily\fontsize{6}{6}\selectfont]
[node_id](node_type){node_description}
[edge_id](parent_node_id, child_node_id){edge_description}
\end{lstlisting}

Nodes are of the following types:

- assistant: message sent by the agent.

- user: message sent by the user.

- api: api call the agent should perform.

\vspace{0.2cm}

Graph construction rules:

- api nodes have outgoing edges with labels

- api nodes are followed by api or assistant nodes

- user nodes are followed by api or assistant nodes

- assistant nodes **can be only followed by** user nodes

- leaf nodes are assistant nodes

\end{tcolorbox}

\begin{tcolorbox}[colback=black!5!white, colframe=black!75!black]
\scriptsize 

Edges connect user nodes to either assistant or api nodes. Only edges from API calls can have descriptions.

The first node should start with an assistant node without any parent node.

For instance, consider the following flow graph:
\begin{lstlisting}[basicstyle=\ttfamily\fontsize{6}{6}\selectfont]
<flow>
[N0](start_message){Greet the customer}
[E0](N0, N1){Didn't receive my order}
[N1](message){Ask customer for order id}
[E2](N1, N2){Gives order id}
[N2](api){get_order_details}
[N3](message){Do you want to cancel or refund the order?}
[E3](N2, N3){Found order}
[N4](message){Tell the user the order wasn't found}
[E4](N2, N4){Order not Found}
[E5](N4, N2){User gives another order id}
[N5](api){cancel_order}
[E6](N3, N5){I want to cancel the order}
[N6](end_message){Order cancelled}
[E7](N5, N6){Success}
[N7](api){refund_order}
[E8](N3, N7){I want a refund}
[N8](end_message){Order refunded}
[E9](N7, N8){Success}
</flow>
\end{lstlisting}
The correct output is:

\begin{lstlisting}[basicstyle=\ttfamily\fontsize{6}{6}\selectfont]
<flow>
[N0](assistant){Greet the customer}
[N1](user){Didn't receive my order}
[E0](N0, N1){}
[N2](assistant){Ask customer for order id}
[E1](N1, N2){}
[N3](user){Gives order id}
[E2](N2, N3){}
[N4](api){get_order_details}
[E3](N3, N4){}
[N5](assistant){Do you want to cancel or refund the order?}
[E4](N4, N5){Found order}
[N6](assistant){Tell the user the order wasn't found}
[E4](N4, N6){Order not Found}
[N7](user){User gives another order id}
[E5](N6, N7){}
[E6](N7, N4){}
[N8](user){I want to cancel the order}
[E7](N5, N8){}
[N9](api){cancel_order}
[E8](N8, N9){}
[N10](assistant){Order cancelled}
[E9](N9, N10){Success}
[N11](user){I want a refund}
[E10](N5, N11){}
[N12](api){refund_order}
[E11](N11, N12){}
[N13](assistant){Your order has been refunded}
[E12](N12, N13){Success}
</flow>
\end{lstlisting}
\end{tcolorbox}

%% file: prompts/conversation_graph_generator_user_prompt.tex
\begin{tcolorbox}[
    colback=black!5!white, 
    colframe=black!75!black, 
    title={User prompt},
]
\begin{lstlisting}[basicstyle=\ttfamily\fontsize{6}{6}\selectfont]
{{ flowgraph }}
\end{lstlisting}

\end{tcolorbox}

%% file: prompts/conversations_generator_system_prompt.tex
\begin{tcolorbox}[colback=black!5!white, colframe=black!75!black, title=System prompt]
\scriptsize 
You will receive a conversation graph with nodes and edges in the following format:

-\texttt{[Ni](assistant)\{message\}}: Agent nodes with the corresponding message.

-\texttt{[Nj](user)\{message\}}: User nodes with the corresponding message.

-\texttt{[Nk](api)\{message\}}: API nodes with the corresponding message.

\vspace{0.2cm}

The graph also has edges with the following format:

-\texttt{[Ei](Ni,Nj)\{\}}: Message Ni happens before Nj.

-\texttt{[Ej](Ni,Nj)\{api\_output\}}: Only applicable when Ni is an API node.

Message Ni happens before Nj and has api outputs api\_output.

The flowchart is given inside <flow></flow>. The initial node is [N1]. The agent is guiding the user throughout the process. Our goal is to generate conversations based on the graph that follow the specified paths, given between <paths></paths>. 
\end{tcolorbox}

\begin{tcolorbox}[colback=black!5!white, colframe=black!75!black]
\scriptsize

For instance, consider the following flow graph:
\vspace{-0.2cm}
\begin{lstlisting}[basicstyle=\ttfamily\fontsize{6}{6}\selectfont]
<flow>
[N1](assistant){Greet the customer}
[N2](user){Didn't receive my order}
[E1](N1, N2){}
[N3](assistant){Ask customer for order id}
[E2](N2, N3){}
[N4](user){Gives order id}
[E3](N3, N4){}
[N5](api){get_order_details}
[E4](N4, N5){}
[N6](assistant){Want to cancel or refund the order?}
[E5](N5, N6){Found order}
[N7](assistant){Tell user the order wasn't found}
[E5](N5, N7){Order not Found}
[N8](user){User gives another order id}
[E6](N7, N8){}
[E7](N8, N5){}
[N9](user){I want to cancel the order}
[E8](N6, N9){}
[N10](api){cancel_order}
[E9](N9, N10){}
[N11](assistant){Order cancelled}
[E10](N10, N11){Success}
[N12](user){I want a refund}
[E11](N6, N12){}
[N13](api){refund_order}
[E12](N12, N13){}
[N14](assistant){Order refunded}
[E13](N13, N14){Success}
</flow>
\end{lstlisting}

And the apis are:
\vspace{-0.2cm}
\begin{lstlisting}[basicstyle=\ttfamily\fontsize{6}{6}\selectfont]
<apis>
[
    {
        "name": "get_order_details",
        "params": [{"order_id": "int"}],
        "output": {'name': 'sent_status', 'type': 'list[dict[str, str]]'}
    }
]
</apis>
\end{lstlisting}

If the given path is: [N1, N2, N3, N4, N5, N7], one possible conversation is the following:

\begin{lstlisting}[basicstyle=\ttfamily\fontsize{6}{6}\selectfont]
[
    {
        "role": "user",
        "content": "I didn't receive my order"
    },
    {
        "role": "assistant",
        "content": "Can you give me the order ID?"
    },
        {
        "role": "user",
        "content": "The order ID is #812"
    },
    {
        "role": "api",
        "content": "get_order_details(order_id=812)"
    },
    {
        "role": "api_output",
        "content": "{"sent_status": [{"item": "Product1", "status": "shipped"}]}"
    },
    {
        "role": "assistant",
        "content": "I couldn't find your order."
    },
]
\end{lstlisting}

Generate the conversation in the format specified above. When making information up, come up with reasonable names and never generic entities like Example1, ProductX, and similar. For example, if talking about products, mention existing products.

Only use the given APIs and make sure all the parameters are defined.
The conversations should follow the following rules:

- After a message with api role always include a message with api\_output role.

- After a message with the assistant role always follow with a message with user role.

- A message with the user role is followed by a message with assistant or api role.

- After a message with a api\_output role always include a message with assistant role.

- The API output should be in the format specified in the API definition. That is always in JSON format.

Note that, even if the node does not exist in the graph, the first message should be a message by the user explaining their problem.
\end{tcolorbox}

%% file: prompts/conversations_generator_user_prompt.tex
\begin{tcolorbox}[
    colback=black!5!white, 
    colframe=black!75!black, 
    title={User prompt},
]
\begin{lstlisting}[basicstyle=\ttfamily\fontsize{6}{6}\selectfont]
{{ conversation_graph }}
<apis>{{ apis }}</apis>
path: {{ path }}
\end{lstlisting}

\end{tcolorbox}

%% file: prompts/conversation_straight_through_generation_system_prompt.tex
\begin{tcolorbox}[colback=black!5!white, colframe=black!75!black, title=System prompt]
\scriptsize 
You are an experienced customer service agent.
You will be given a procedure enclosed by <procedure></procedure> and a list of apis that can used enclosed by <apis></apis>.
Your goal is to generate conversations between an agent and a customer that could be solved used the given procedure and apis.

For instance, consider the following procedure:
\begin{lstlisting}[basicstyle=\ttfamily\fontsize{6}{6}\selectfont]
<procedure>
# Handling a Customer Who Didn't Receive Their Order

Start Interaction:
1.1. Greet the customer courteously.

Identify the Issue:
2.1. Confirm the customer didn't receive the order.

Obtain Order Information:
3.1. Ask the customer to provide their order ID along with the email address or phone number associated with the order.

Retrieve Order Details:
4.1. If the customer provides the order ID and email address:
- Use the company's API to retrieve order details by email.
4.2. If the customer provides the order ID and phone number:
- Use the company's API to retrieve order details by phone number.

Check if Order is Found:
5.1. If the order is found, proceed to Step 6.
5.2. If the order is not found:
- Inform the customer that the order wasn't found.
- Ask the customer to provide the correct email or phone number and order ID.
- Repeat Step 3 based on the new information.

Determine Customer's Request:
6.1. Ask the customer if they would like to cancel the order or request a refund.

Processing Customer's Request:
7.1. Cancellation:
- If the customer wants to cancel the order:
- Use the company's API to cancel the order.
- Upon successful cancellation, inform the customer that the order has been cancelled.
7.2. Refund:
- If the customer wants a refund:
- Use the company's API to process the refund.
- Upon successful refund, inform the customer that the order has been refunded.

End Interaction:
8.1. Conclude by thanking the customer for their patience and confirming resolution.
\end{lstlisting}

And the apis are:
\begin{lstlisting}[basicstyle=\ttfamily\fontsize{6}{6}\selectfont]
<apis>
[
    {
        "name": "get_order_details",
        "params": [{"order_id": "int"}],
        "output": "bool"
    }
]
</apis>
\end{lstlisting}

One possible conversation is the following:
\begin{lstlisting}[basicstyle=\ttfamily\fontsize{6}{6}\selectfont]
[
    {
        "role": "assistant",
        "content": "Hello, how can I assist you?"
    },
    {
        "role": "user",
        "content": "I didn't receive my order"
    },
    {
        "role": "assistant",
        "content": "Can you give me the order ID?"
    },
    {
        "role": "user",
        "content": "The order ID is #812"
    },
    {
        "role": "api",
        "content": "get_order_details(order_id=812)"
    },
    {
        "role": "api_output",
        "content": "False"
    },
    {
        "role": "assistant",
        "content": "I'm sorry but I couldn't find your order."
    },
]
\end{lstlisting}

\end{tcolorbox}

\begin{tcolorbox}[colback=black!5!white, colframe=black!75!black]
\scriptsize

Generate the conversation in the format specified above. When making information up, come up with reasonable
names and never generic entities like Example1, ProductX, and similar. For example, if talking about products,
mention existing products.

Only use the given APIs and make sure all the parameters are defined.
The conversations should follow the following rules:
\begin{lstlisting}[basicstyle=\ttfamily\fontsize{6}{6}\selectfont]
- After a message with api role always include a message with api\_output role.
- After a message with the assistant role always follow with a message with user role.
- A message with the user role is followed by a message with assistant or api role.
- After a message with a api\_output role always include a message with assistant role.
\end{lstlisting}

Note that, even if the node does not exist in the graph, the first message should be a message by the user explaining their problem.

\end{tcolorbox}

%% file: prompts/conversation_straight_through_generation_user_prompt.tex
\begin{tcolorbox}[colback=black!5!white, colframe=black!75!black, title=User prompt]
\scriptsize 
\begin{lstlisting}[basicstyle=\ttfamily\fontsize{6}{6}\selectfont]
<procedure>{{ procedure }}</procedure>
<apis>{{ apis }}</apis>
\end{lstlisting}
\end{tcolorbox}

%% file: prompts/agent_system_prompt.tex
\begin{tcolorbox}[colback=black!5!white, colframe=black!75!black, title=System prompt]
\scriptsize
You are a customer support agent with the goal of answering user requests.
You will be given the following information:
\begin{lstlisting}[basicstyle=\ttfamily\fontsize{6}{6}\selectfont]
- conversation: Messages exchanged between the end user and you, and the executed actions with their 
outputs.
\end{lstlisting}

This is the procedure you know about:
\begin{lstlisting}[basicstyle=\ttfamily\fontsize{6}{6}\selectfont]
<procedure>
{{ procedure }}
</procedure>
\end{lstlisting}
You only know answers about this procedure! It is critical that you do not come up with any data nor instructions that are not contained in the procedure.

This is the list of available actions.
\begin{lstlisting}[basicstyle=\ttfamily\fontsize{6}{6}\selectfont]
<actions>
{{ available_actions }}
</actions>
\end{lstlisting}

Sometimes your action might be simply to reply to an end user, other times you will need to call an action that performs an operation and/or retrieves necessary data.
Some actions require information/parameters in order to be callable. If you do not have the necessary information available in the context, YOU MUST ASK FOR IT AND CANNOT SUGGEST THE ACTION.
Make sure that you follow the directives in the procedure before suggesting a relevant action. For instance, some actions have consequences and might require user confirmation before being executed, if stated in the procedure. If this is the case, suggest a reply that asks confirmation from the end user.
Make sure that the information that you are using properly matches the context (e.g., the user might give a phone number that does not match what is shown in the context, which contains the output of actions.)

You MUST reply with a JSON object as follows:
\begin{lstlisting}[basicstyle=\ttfamily\fontsize{6}{6}\selectfont]
{
    'type': name of the function to call,
    'parameters': parameters to pass to the function,
}
\end{lstlisting}
\end{tcolorbox}

%% file: prompts/agent_user_prompt.tex
\begin{tcolorbox}[colback=black!5!white, colframe=black!75!black, title=User prompt]
\scriptsize
\begin{lstlisting}[basicstyle=\ttfamily\fontsize{6}{6}\selectfont]
<conversation>
{{ conversation }}
</conversation>
\end{lstlisting}
\end{tcolorbox}